\documentclass{article}

\usepackage[preprint]{neurips_2022}

\usepackage[utf8]{inputenc} 
\usepackage[T1]{fontenc}    
\usepackage{hyperref}       
\usepackage{url}            
\usepackage{booktabs}       
\usepackage{amsfonts}       
\usepackage{nicefrac}       
\usepackage{microtype}      
\usepackage{xcolor}         
\usepackage{microtype}
\usepackage{graphicx}
\usepackage{subfigure}
\usepackage{booktabs} 
\usepackage{bm}
\usepackage{graphicx}
\usepackage{ulem}
\usepackage{amsmath}
\usepackage{amssymb}

\usepackage{wrapfig}
\usepackage{algorithm}
\usepackage{algorithmicx}
\usepackage{algpseudocode}

\title{Ensembling improves stability and power of feature selection for deep learning models}

\author{Prashnna K Gyawali\qquad Xiaoxia Liu\qquad James Zou$^{*}$ \qquad Zihuai He$^{*}$\\ \\ Stanford University}

\begin{document}
\maketitle

\begin{abstract}
With the growing adoption of deep learning models in different real-world domains, including computational biology, it is often necessary to understand which data features are essential for the model's decision. Despite extensive recent efforts to define different feature importance metrics for deep learning models, we identified that inherent stochasticity in the design and training of deep learning models makes commonly used feature importance scores unstable. This results in varied explanations or selections of different features across different runs of the model. We demonstrate how the signal strength of features and correlation among features directly contribute to this instability. To address this instability, we explore the ensembling of feature importance scores of models across different epochs and find that this simple approach can substantially address this issue. For example, we consider knockoff inference as they allow feature selection with statistical guarantees. We discover considerable variability in selected features in different epochs of deep learning training, and the best selection of features doesn't necessarily occur at the lowest validation loss, the conventional approach to determine the best model. As such, we present a framework to combine the feature importance of trained models across different hyperparameter settings and epochs, and instead of selecting features from one best model, we perform an ensemble of feature importance scores from numerous good models. Across the range of experiments in simulated and various real-world datasets, we demonstrate that the proposed framework consistently improves the power of feature selection.
\end{abstract}

\section{Introduction}
Machine learning (ML) algorithms, especially deep learning models, are being used extensively and making important decisions in real-world domains, including medicine and the biological domain. 
These algorithms are often required to be interpretable as this allows us to understand which input features are used to make the decision. For some applications, knowing which features are selected enhances trust in the model, while for other applications, selected features may help in scientific discovery, e.g., drug discovery problems. Overall, interpretability and explainability are critical with the growing usage of ML algorithms. However, although recent efforts have given rise to various interpretability methods \cite{sundararajan2017axiomatic, shrikumar2017learning, binder2016layer, kim2018interpretability} and have explained why an algorithm made certain decisions, these interpretations remain fragile \cite{ghorbani2019interpretation}, leading to different explanations for the same data instance and, eventually, contributing toward mistrust in the models. 

The primary reason for such varied explanations can be attributed to stochasticity in the design and training of deep neural networks (DNN). Some examples could be model initialization, dropouts, stochastic gradient descent, etc. For instance, the unstable behavior of DNN from varying only the initialization of the network weights has been reported before \cite{mehrer2020individual}. 
Further, with the objective function of neural networks being highly non-convex, the resulting model may end up in different local minima. Although such local minima result in models with similar generalization performance (as measured with validation loss), these result in varied explanations or feature importance scores for the same data.
This instability in feature attribution is aggravated by the dataset having low signal amplitude features or high correlation between features which is quite common in real-world data analysis. More importantly, this instability in feature importance score directly impacts the stability of selected features. 

In this work, we first demonstrate this issue of instability in feature importance and feature selection for standard benchmarking datasets and interpretability metrics. We also provide evidence of how data properties like signal strength and correlation aggravate instability. We then demonstrate how simple averaging of feature importance scores from models at different training epochs helps address this instability. Motivated by the abilities of such averaging, we propose a framework for stabilizing the feature importance and feature selection from the deep neural network. Our proposed framework first perform hyperparameter optimization of deep learning models. Then, instead of the conventional practice of selecting a single best model, we find out numerous \textit{good} models and create an ensemble of their feature importance score, which, as we show later, will help select robust features. For determining good models, we consider two strategies. First, we propose using top-performing models as determined by cross-validation (CV) loss. In the second strategy, we propose statistical leveraging to find the influential models for feature importance. 
In this work, we consider a knockoff framework for feature selection as they choose features with statistical guarantees. 
Across a range of experiments in simulation settings and real-world feature selection problems, we find that the existing approach 
of selecting features from the best model across different hyperparameter settings and epochs doesn't necessarily result in stable or improved feature selection. 
Instead, we achieve stable and improved feature selection with the presented framework. Overall, our contributions are as follows:
\begin{itemize}
    \item We demonstrate the instability in DNN interpretations for widely used interpretability metrics (Grad, DeepLift and Lime) across two benchmarking datasets (MNIST and CIFAR-10).
    \item We propose a framework to create an ensemble of feature importance scores obtained from the training paths of the deep neural network to stabilize the feature importance score from deep neural networks.
    \item We demonstrate the applicability of such an ensemble in the task of feature selection with knockoff inference. 
    \item Across the simulation studies and three real-data applications for feature selection, we demonstrate the efficacy of the proposed framework to improve both stability and the power of feature selection for deep learning models.
\end{itemize}

\section{Related works}
In recent times, there have been works that carefully studied the fragility in neural networks interpretation \cite{ghorbani2019interpretation,slack2020fooling}. These works demonstrate that explanation approaches are fragile to adversarial perturbations where perceptively indistinguishable inputs can have very different interpretations, despite assigning the same predicted label. Although our work is also about the instability in interpretations of the neural network, unlike them, we study this problem without relying on adversarial inputs. Further, we focus on the impact of this instability on the downstream application of feature selection, which hasn't been considered before.

The primary strategy in our framework, \textit{i.e.,} ensemble of feature importance score from models in different training stages has some similarities with the recent works in deep learning generalization \cite{li2022trainable, izmailov2018averaging}. These works have studied the model's weight averaging as an alternative to improve generalization. However, unlike these works, we propose to form an ensemble of feature importance scores obtained from individual model weights at different stages of the deep learning training, and most importantly, unlike all previous works, we used such an ensemble to improve the stability and power of feature selection.

Feature selection (or variable selection) has been extensively studied in machine learning and statistics \cite{saeys2007review,mares2016combining}. The selection of features while controlling false discovery is an attractive property, and there exist different feature selection methods that provide such statistical guarantees \cite{meinshausen2010stability, barber2015controlling}.
Although our framework applies to any such feature selection method, we consider knockoff inference for feature selection \cite{barber2015controlling}.

Since we consider knockoff inference to demonstrate how our proposal helps to stabilize and improve feature selection in deep learning, our work is also related to research on the intersection of deep learning and knockoff inference. 
Despite being model-free and selecting features with statistical guarantees, the usage of knockoff inference with deep learning is quite limited \cite{lu2018deeppink, zhu2021deeplink}. 
Our work is complementary to these works as we demonstrate the instability issues in feature selection with the help of these works, and our proposed framework improves the power of feature selection compared to such works. Furthermore, within the knockoff framework, the idea of constructing multiple knockoffs has been studied in detail to improve feature selection \cite{he2021identification, gimenez2019improving}. Our work utilizes single and multiple knockoffs depending on the dataset's complexity and demonstrates the utility of the presented framework for both scenarios. 


\begin{figure*}[!t]
\begin{center}
    \includegraphics[width=\linewidth]{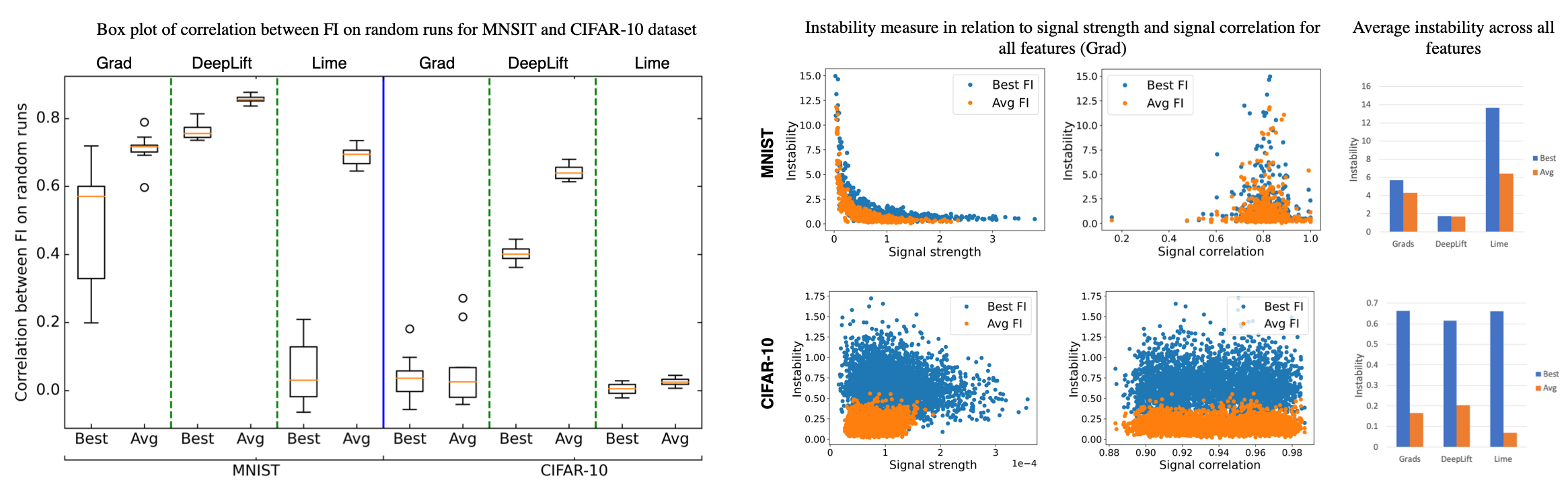}
  \end{center}
  \vspace{-0.3cm}
\caption{(Left) Box plot of correlation between feature importance scores between models trained separately with five random initialization. This analysis considers two datasets (MNIST and CIFAR-10) and three different feature importance metrics (Grad, DeepLift, and Lime). The \textit{Best} feature importance score (obtained from the model with the lowest validation loss) is compared with the \textit{Avg} feature importance score obtained from the ensemble of feature importance scores from all training epochs. (Middle) Instability of individual features across different randomly initialized models in relation to their signal strength and signal correlation with all other features. (Right) Average instability across all the features for three feature importance measures considered in this study.}
\vspace{-0.5cm}
\label{fig:randomness_analysis}
\end{figure*}

\section{Instability in models' interpretability}
In this section, we explore instability in the interpretations of deep neural networks. Toward this we consider two benchmarking datasets: MNIST \cite{lecun1998gradient} and CIFAR-10 \cite{krizhevsky2009learning}. We train the models for the standard image classification task and record their feature importance (FI) score (details in Appendix A). In particular, for a given dataset, we train a deep neural network five times with random initialization and record the FI for the best model (as defined by the lowest validation loss in the training epochs) in each run. 
We randomly sample (stratified sampling to cover all the classes in the dataset) images from the test dataset to calculate the FI and average their absolute values to obtain a single FI score $Z$.
For FI, we consider three separate and widely considered feature importance metrics: \textit{Grad}, gradients with respect to inputs, \textit{DeepLift}, a back-propagation-based approach that attributes a change to inputs based on the differences between the inputs and corresponding references (or baselines) for non-linear activations \cite{shrikumar2017learning}, and \textit{Lime}, an interpretability method that trains an interpretable surrogate model by sampling data points around a specified input example and using model evaluations at these points to train a simpler interpretable `surrogate' model \cite{ribeiro2016should}. Ideally, to provide robust interpretability, we expect the correlation between different $Z$ from these randomly initialized models to be close to 1. However, as we see in Fig. \ref{fig:randomness_analysis} (left), for both datasets and across all the FI metrics, the best FI score between the same models trained with different random initialization are not correlated.
Instead, when we create an ensemble from $Z$ obtained across $E$ epochs, \textit{i.e.,} $Z_{\textrm{avg}}  =  \frac{1}{E} \sum^{E}_{i = 1} Z_{i}$,
we found that such an ensemble helps in stabilizing the interpretations with increased correlation between these random runs. As we can see in Fig. \ref{fig:randomness_analysis} (left), for MNIST, the ensemble (represented by Avg) increases the correlation significantly, and for CIFAR-10, although the ensemble wasn't able to increase the correlation by a considerable margin, it was still doing better compared to the best model.


We then proceed to understand how the individual features in the dataset is contributing toward this instability issue. Toward this we consider two attributes of the dataset: signal strength of the individual features and correlation among the features. First, for each feature point $j \in \{1, .., p\}$ in the dataset $\mathbf{X} \in \mathbb{R}^{p}$ ($p = 784$ for MNIST and $p = 3072$ for CIFAR-10), we calculate the instability metric for each $Z^{j}$ ($j \in [1, .., p]$) as $\frac{\sigma(Z^{j}_{[1:5]})}{\mu(Z^{j}_{[1:5]})}$,
where $\sigma$ and $\mu$ are, respectively, the standard deviation and the average across given values, and [1:5] represents the result obtained from five randomly initialized experiments.
Second, we define signal strength for each feature as $\mu(Z^{j}_{[1:5]})$ and for correlation between features, we calculate maximum correlation of given feature dimension $j$ to all other $k$ features \textit{i.e.,} $\textrm{max(Corr}(Z_j, Z_k))$ for $k \in [p]; k \neq j $, where $\textrm{Corr}$ is the Pearson correlation coefficient. We provide the result of this analysis in Fig. \ref{fig:randomness_analysis} (middle) with Grad as the FI score. As we can see for both datasets, the lower signal strength of the feature and higher correlation among features manifests unstable behavior in feature importance. This means that besides the inherent stochasticity in the training of the deep neural network, the properties of individual features like signal strength and correlation among features also create instability in the feature importance, making interpretation of the results unreliable. With the ensembling of feature importance score (Avg FI), the instability is reduced.
We also plot the bar diagram in Fig. \ref{fig:randomness_analysis} (right) for all the feature importance metrics and see that the average instability across all the features is reduced with the ensembling approach compared to the conventional approach of using the best model from the validation loss.

We consider this instability analysis as the first contribution of our paper, which points out that the standard training setup of deep neural networks and their corresponding interpretations enhance the mistrust in the models. With the ensemble, we can provide stable interpretations. However, interpreting models with stable but inconsequential features is also not our goal. As such, we aim to investigate if an ensemble of feature importance scores also helps select important features. As such, in the rest of this paper, we study feature selection with deep learning. 

\begin{figure*}[t]
\begin{center}
    \includegraphics[scale=0.35]{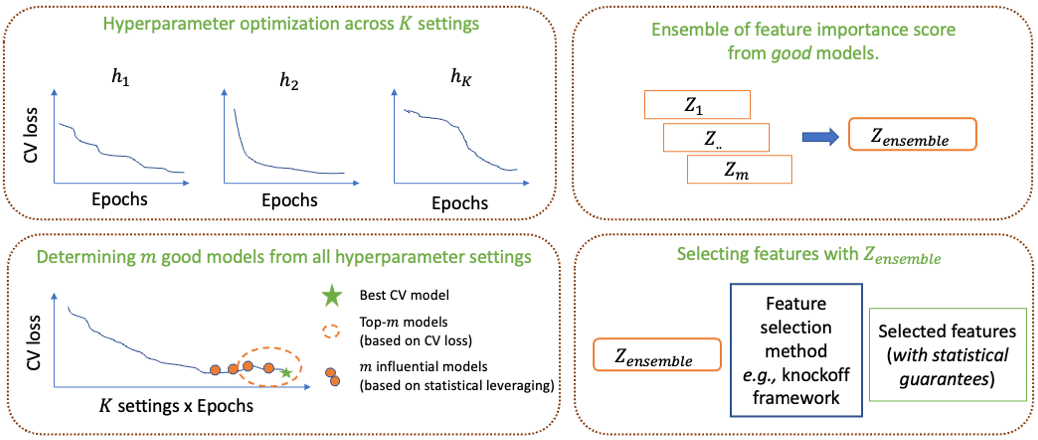}
  \end{center}
  \vspace{-0.3cm}
\caption{Schematic diagram of the proposed framework. 
The framework performs hyperparameters optimization across $K$ different settings, then selects $m$ good models instead of choosing a single best model. Third, from the $m$ selected models, it creates the ensemble feature importance scores $Z_{\textrm{ensemble}}$, which is used for the feature selection to select the stable and important features. 
}
\label{fig:schematic}
\end{figure*}

\section{Ensemble framework to improve stability and power of feature selection }
In this section, we present the framework based on feature importance ensembling to improve the stability and power of feature selection for deep learning models. As outlined in Fig \ref{fig:schematic}, our framework comprises four steps: hyperparameters optimization of deep learning models, determining the \textit{good} models, an ensemble of feature importance score, and feature selection with the ensemble of feature importance scores. For feature selection, we consider a knockoff framework as it allows
a way to evaluate power while controlling for false discoveries.


\subsection{Problem formulation}
We consider dataset $\mathcal{D} = \{\mathbf{X}^{i}, \mathbf{Y}^{i} \}^{N}_{i=1}$. Here, $\mathbf{X} \in \mathbb{R}^{p}$ represents input data and $\mathbf{Y}$ represents its corresponding outcome (or labels). We are interested in discovering a set of features $\Tilde{P} \subset \{X_1, .. X_p \}$ that holds causal link of input data and the output labels. Towards discovering a set of features, we consider knockoff framework and create knockoff copies of original data $\Tilde{\mathbf{X}}$, where $\Tilde{\mathbf{X}} \in \mathbb{R}^{p}$ (for single knockoff) and $\Tilde{\mathbf{X}} \in \mathbb{R}^{m * p}$ (for $m$ multiple knockoff). 

\subsection{Background: Knockoff Framework}
Knockoff framework \cite{barber2015controlling, candes2018panning} has recently emerged as a powerful and flexible method to perform controlled variable selection. Informally, given a set of features ($X_1, .., X_p$) and an outcome of interest $Y$, knockoffs allow one to leverage almost any learning method to discover relationships between the features and the outcome. Notably, knockoffs exactly control the expected proportion of false positives in finite samples provided that the distribution of the features $\mathbf{Y}$ is known while assuming nothing about the conditional distribution $Y | X$. Toward this, the knockoff framework constructs synthetic variables, which mimic the correlation structure of the original features. By contrasting the original and knockoff data, the knockoff-based method allows the selection of important features related to the outcome of interest while controlling the false discovery rate (FDR). We provide more details in Appendix B.

\subsection{Hyperparameters optimization in deep learning}
In order to find the important features from $\mathbf{X}$, we consider standard setup in deep learning training for finding the best model. We consider $\{h_k\}^{K}_{k=1}$ as the $K$-different hyperparameter settings for training deep learning models. These hyperparamter settings can include the choice of $\lambda$ values for $L1$ penalty, depth of neural network, choice of activation functions, etc. For each hyperparameter $h_k$, we train the model for the fixed $E$ epochs, and record the corresponding CV loss $l$ and feature importance score $Z$ at each epoch. For example, $l^{e}_{k}$ refers to the CV loss, and $Z^{e}_{k}$ refers to the feature importance score at epoch $e$ during $k$ hyperparameter training. 
The conventional approach in deep learning is to find $k^{\textrm{th}}$ setting and $e^{\textrm{th}}$ epoch which produces the lowest CV loss and use its corresponding feature importance score for interpreting the models and selecting the features. 
We refer such model as best CV model, which is defined as $Z^{e}_{k} [\underset{k, e}{\mathrm{argmin}} \quad l^{e}_{k}]$.
In this work, we consider all $n$ ($n = K \times E)$ models in the training regime to determine the $m$ ($m \leq n$) \textit{good} models for finding the stable features.

\subsection{Determining \textit{good} models}
Since the loss landscape of neural network training is non-convex with many solutions in different loss basins, in this work, we argue that instead of selecting a single best model, we should also look for other \textit{good} models that have captured the relationship between $X$ and $Y$, and hence can provide robust and stable feature selection. We propose two strategies to find the good models: 1) select top-$m$ $Z$ based on the lowest CV loss, and 2) select $m$ influential $Z$ using the statistical leveraging.

In the first strategy, we combine CV loss from all the $K$ settings and all $E$ epochs within each setting. We then sort these $n$ CV losses in descending order and select the top-$m$ $Z$ representing the models with the lowest CV loss. Since we arrange all the settings together, these $Z_{[1:m]}$ may represent feature importance scores for different settings and different epochs within each setting.  

In the second strategy, we propose \textit{statistical leveraging} to find the influential $Z$ within the deep learning training regime. 
The statistical leverage scores have been extensively used (primarily within classical regression diagnostics) to determine an influential observation \cite{izmailov2018averaging}. 
To the best of our knowledge, our work is the first to consider this classical statistical approach to determine the significant feature importance score within the neural network training regime. 
Toward this, we combine all $n$ feature importance scores and calculate each observation's statistical leverage. 
The obtained statistical leverage of each observation is used as sampling weights to obtain $Z_{[1:m]}$, $m$ influential $Z$. 
We provide the algorithm in Appendix C. 

\begin{figure}[t]
\centering
\includegraphics[scale=0.30]{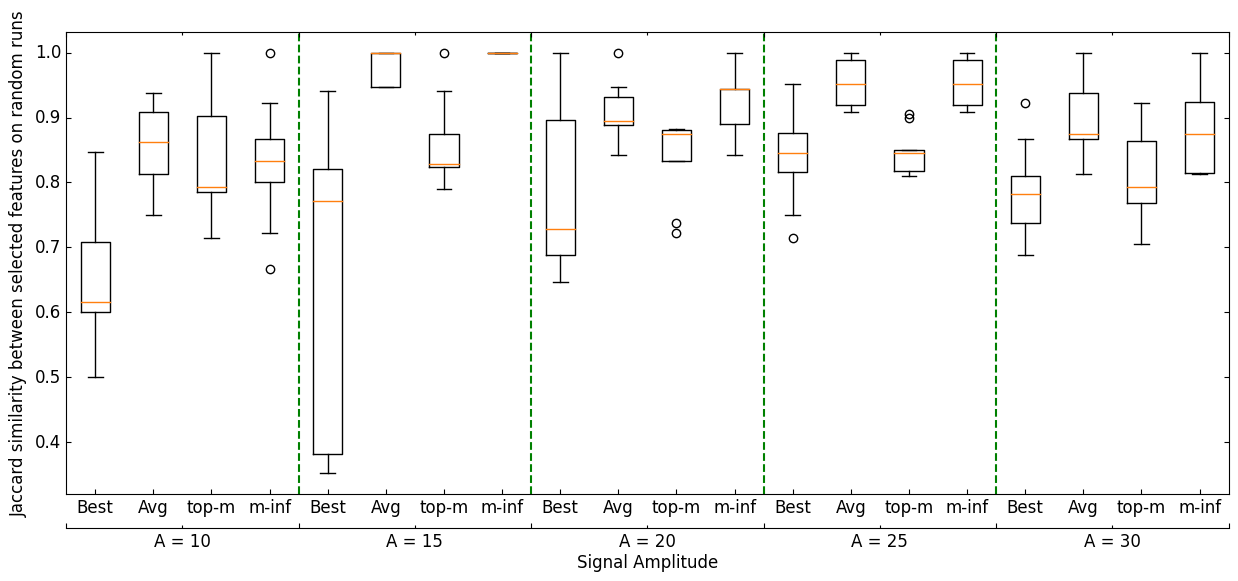}
\vspace{-0.4cm}
\caption{Jaccard similarity between selected features on five random runs for five different simulated datasets. The ensemble models (\textit{Avg}, \textit{top-$m$}, and \textit{$m$-inf}) is compared with the best CV model. 
}
\vspace{-0.4cm}
 \label{fig:boxPlot_jaccard}
\end{figure}

\subsection{Ensemble of feature importance score from good models}
In this step of the framework, we combine $m$ feature importance scores obtained from the previous step to create the ensemble of feature importance score $Z_{\textrm{ensemble}}$, \textit{i.e., } $\frac{1}{m} \sum^{m}_{i = 1} Z_{[1:m]}$.
Note that $Z_{\textrm{avg}}$ is different from $Z_{\textrm{ensemble}}$ as it is the average of all $n$ $Z$ in the training regime.

\subsection{Knockoff filter for feature selection}
We combine the original features $Z_j$ and knockoff features $Z_{j+(m_{k}*p)}$ (for ensemble feature importance $Z_{\textrm{ensemble}}$) into a feature statistic $W_j = f_s (Z_j, Z_{j+(m_{k}*p)})$, where $f_s$ is an antisymmetric function such that $f_s(x, y) = - f_s(y,s)$, and $m_k$ is the number of knockoff copies considered in the multiple knockoff settings ($m_{k}$ = 1 for single knockoff). 
For single knockoff case, if $Z$ are absolute coefficients (or absolute values of gradients for deep neural networks), we can set $f_s (Z_j, Z_{j+p}) = Z_j - Z_{j+p}$. 
Using feature statistics $W_j$, we then apply a knockoff filter to calculate the threshold $T$ for the target FDR level $q$. 
We provide details for calculating feature statistic $W_j$ and knockoff filter in Appendix B.
With the calculation of threshold $T$, we select $\Tilde{P} = \{j: W_j \geq T \}$. This guarantees the $\Tilde{P}$ selected features from $Z_{\textrm{ensemble}}$ are FDR control at level $q$.


\section{Experiments}
For all our feature selection experiments, we constructed neural network architectures based on DeepLink \cite{zhu2021deeplink}.
We generate a single knockoff for simulation studies, and for real-world experiments, we generate multiple knockoffs. For both simulated and real-world experiments, the target FDR $q$ is set to 0.2.
Considering the feature selection task, we followed the standard practice of cross-validation utilizing all the samples in the dataset. For each hyperparameter setting, we use 5-fold cross-validation to train the models, record the average CV loss, and re-train the model using all the datasets. Finally, we use the CV losses and the re-trained models to determine the best and $m$ good models. 
We consider $\lambda$ values (for L1 regularization) and the depth of neural network as our hyperparameters for simulation studies and only $\lambda$ values for real-world experiments.

\subsection{Simulation Studies}
We consider the following linear factor model for generating $n$ data points with $p$ features:
\begin{equation}
    \mathbf{x}_i = \Lambda \cdot \mathbf{f}_i + \epsilon_i
    \label{eq:sim_data}
\end{equation}
where $\mathbf{f}_i \in \mathbb{R}^{r}$ is the vector of latent factors, $\Lambda \in \mathbb{R}^{r \times p}$ is the factor loading matrix and $\epsilon \in \mathbb{R}^{n}$ is the noise parameter. The entries of $\mathbf{f}_i$, $\Lambda$ and $\epsilon_i$ are all sampled independently from the standard normal distribution $N(0,1)$. The response vector $\mathbf{y} = (y_1, ..y_n)^T$ is simulated as $\mathbf{y} = \mathbf{x} \cdot \beta  + \epsilon$,
where $\mathbf{x} \in \mathbb{R}^{n\times p}$ is the data matrix generated from Eq. \ref{eq:sim_data}, $\mathbf{\beta} = (\beta_1, .., \beta_p)^T \in \mathbb{R}^p$ is the coefficient vector, and $\epsilon \in \mathbb{R}^n$ is a vector of noise. To simulate the coefficient vector $\beta$, we first randomly choose $s$ true signal locations and then set the $\beta_j = \textrm{sample}([A, -A])$ with equal probability, where $A$ is a positive amplitude values. The remaining $p - s$ elements of $\beta$ are set to zero. For all simulation studies, the sample size $n$ is set to 1,000, $p$ is set to 500, $r$ is set to 3, true signal size $s$ is set to 25. We varied the amplitude values $A = \{10, 15, 20, 25, 30\}$ and thus, experimented with the resulting five different data settings. For each such data setting, we created 100 replicates to explore the performance of the presented ensembling framework. 

\begin{figure*}[t]
\begin{center}
    \includegraphics[width=\linewidth]{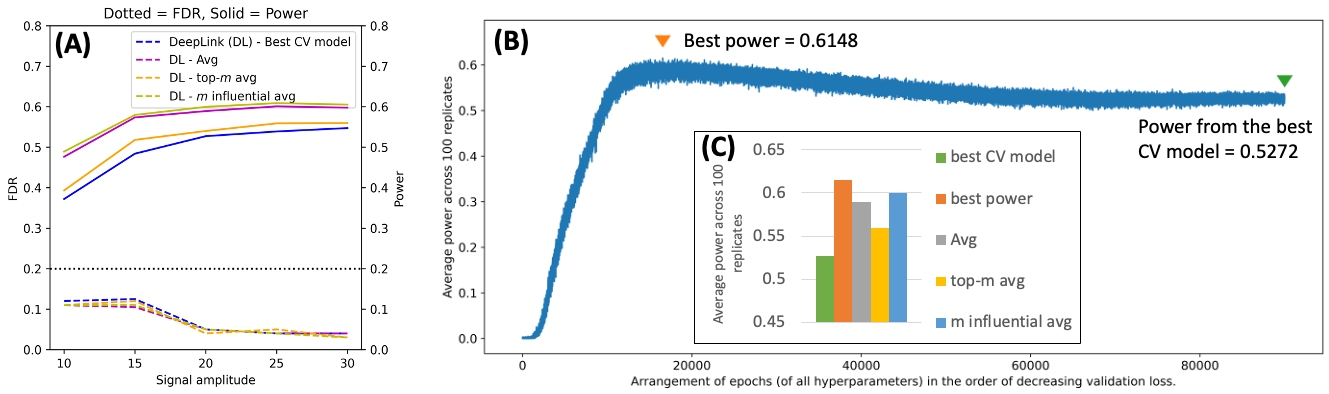}
  \end{center}
  \vspace{-0.3cm}
\caption{(A) Average power and FDR across 100 replicates achieved by the best CV model and the ensembling approaches for five different amplitude values. The black dashed line indicates the target FDR level. (B) Power vs. CV loss for all 90,000 settings considered in the simulation study for $A = 20$ demonstrates the best CV model doesn't necessarily achieve the best power. (C) With ensembling, the power can be improved compared to the best CV model. }
  \vspace{-0.3cm}
\label{fig:sim_main_result}
\end{figure*}

We next provide the details of the training setup of each experiment. We consider neural network architecture with three different depths (depth = 1, 2, and 3) and 100 different lambda values (see Appendix A for more details). For each hyperparameter setting, we train our model for 300 epochs and optimize using Adam with a constant learning rate of 0.001. This resulted in total of 9,000,000 models (100 data replicates $\times$  300 epochs $\times$ 100 $\lambda$ values $\times$ 3 depths) for analysis. We report all our results as the average of 100 replicates reducing our analysis space to 90,000, where we set $m$ as 500 for finding good models.
We use top-$m$ models based on CV loss to determine $Z_{\textrm{ensemble}}$, represented by top-$m$ avg and use statistical leveraging presented in Algorithm 1 to determine $m$ models for calculating $Z_{\textrm{ensemble}}$, represented by $m$ influential avg (or $m$-inf).

\subsubsection{Results} 
We first demonstrate the instability issue of deep learning models in feature selection, similar to instability in feature importance as demonstrated before. For this analysis, we randomly picked one replicate from the simulated datasets and one hyperparameter setting for each signal amplitude. We train the given deep learning model five times, varying only the weight initialization (with random seeds). We then calculated the Jaccard similarity coefficient to calculate the similarity of the selected features. We present the results in Fig. \ref{fig:boxPlot_jaccard} where we observe the variations between the selected features for all amplitude values considered in this work. The result also demonstrates the reduced variations in selected features with averaging (Avg) and two strategies in our presented framework (top-$m$ avg and $m$-inf avg). 
Also, note that the variability is high for data settings with low signal amplitude, 
which is consistent with our earlier observation. 

We then calculate the power of feature selection \textit{i.e.,} the quality of selected features via the proposed ensemble framework compared to the best CV model. Here, we consider all datasets (100 replicates) and all hyperparameter settings. We present the results in Fig. \ref{fig:sim_main_result}. First, in Fig. \ref{fig:sim_main_result} (A), we present the obtained power and FDR for all the amplitude values together, demonstrating that the proposed framework achieves better power (with FDR under control) compared to the standard setup of using the best CV model for feature selection.
Second, we try to understand how ensembling helps us achieve better power, and for that, we looked at the obtained power at each training epoch for all the hyperparameter settings. We present this analysis in \ref{fig:sim_main_result} (B), where we plot the obtained power in decreasing order of validation loss such that the right-most point represents the result from the best model.
As we can see, the power (0.5272) from the conventional approach is much lesser than the best power (0.6148) in the hyper-parameter space. With ensembling approach, as shown in Fig. \ref{fig:sim_main_result} (C), we can improve the power up to 0.5588 (via top-$m$ avg) and up to 0.5996 (via $m$ influential avg).  Overall, these results demonstrate that ensembling not only helps in stabilizing the feature importance score and feature selection but also helps improve the power of feature selection.



\begin{table}
	\begin{minipage}{0.5\linewidth}
		\centering
		\small{
        \begin{tabular}[t]{|c|c|c|c|}\hline
          Dataset & Proteomics & Breast & Pima \\
            & & Cancer & Diabetes \\
            \hline
            Best CV model & 46 & 3 & 6 \\
            Avg & 8 & 12 & 5\\ 
           \hline
            top-$m$ avg & 60 & 16 & \textbf{7} \\
            $m$ influential avg & \textbf{61} & \textbf{17} & 6 \\
            \hline 
        \end{tabular}
        }
	\end{minipage}\hfill
	\begin{minipage}{0.42\linewidth}
		\centering
		\includegraphics[scale=0.38]{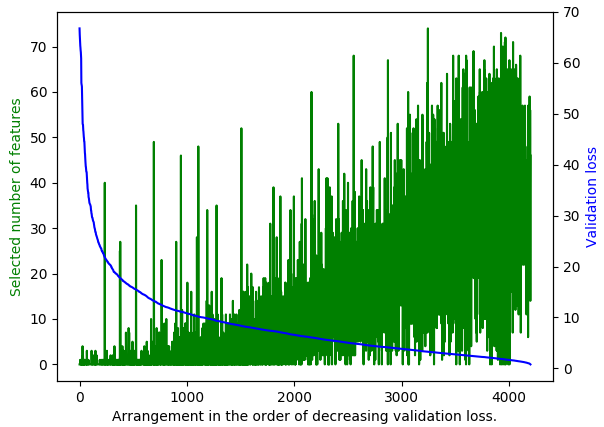}
	\end{minipage}
	\caption{(left) Number of features selected with FDR control ($q$ = 0.2) for different real-world datasets. The two strategies of the proposed framework are below the dotted line. (right) Inversely sorted (high-to-low) CV loss vs. selected number of features for Proteomics dataset demonstrating high feature selection variability in training paths of DNN.}
	\label{tab:real_results}
	\vspace{-0.5cm}
\end{table}

\subsection{Real-data experiments}
For real-data experiments, we consider three different biological datasets: the Proteomics dataset, Breast Cancer dataset, and Pima Diabetes dataset (see Appendix D for more details). Proteomics dataset ($n = 675$, $p = 7596$) is used for binary classification between healthy control and Alzheimer's Disease. Breast Cancer dataset ($n = 569$, $p = 30$) is used for binary classification to separate between cancer and non-cancer individuals. Finally, Pima Diabetes ($n = 768$, $p = 8$) is used to diagnostically predict whether or not a patient has diabetes. Since single knockoff can have limitations in terms of detection threshold and stability, which can be critical for real-world datasets, we considered multiple knockoffs for all three real-data experiments. 

For training the DeepLink models, we considered 20 different $\lambda$ parameters as hyperparameters for all three datasets. We trained them for $E$ epochs, where $E = 200$ for Proteomics, $E = 100$ for Cancer, and $E = 50$ for Diabetes, and optimized all of them using Adam with a constant learning rate of 0.001. This resulted in a total of 4,000 models for Proteomics, 2,000 for Cancer, and 1,000 for Diabetes. To obtain the $m$ good models, we set $m$ to 100, and acknowledging the numerous bad models in the training paths (unlike simulation setting where DNN training is quite stable, resulting in numerous good models), we search for influential models within the top-25 percentile of the lowest CV loss. For $m$-top models, we used the same strategy as before. 


\subsubsection{Results} We present our main results in Table \ref{tab:real_results} (left) where we can see that the proposed framework, similar to the simulation studies, can select more features compared to the conventional approach of selecting features from the best CV model. Although we don't have ground-truth features to compare the importance of these selected features, the higher number of features represents more power in feature selection as they are selected by controlling the FDR ($q = 0.2$).
To further understand how ensemble approaches are helping to select more features, we want to highlight the instability issues in these experiments as well. In Table \ref{tab:real_results} (right), we present the CV loss and corresponding selected features for each training epoch for the Proteomics dataset, where we can see that despite the smooth decrease in the validation loss (blue curve), there is big variability in the selected features (represented by green lines). The proposed framework, via an ensemble of good models, can reduce this variability and provide more power in the feature selection. Also, note that the presence of such bad models has directly impacted the ability to select a high number of controlled features via simple averaging (Avg) in Table \ref{tab:real_results}. The Best model has more power than the Avg model in two out of three cases. However, both strategies in the presented framework produce better power in all three datasets. Overall, these results demonstrate the practical utility of the presented framework on real data applications.

\section{Conclusion}
This paper demonstrates the instability issues in interpreting deep neural networks and their effect on feature selection. We first illustrate the instability in feature importance calculated via standard interpretability measures on widely used benchmarking datasets. We also showed how simple averaging of feature importance across the training paths helps alleviate this issue. 
We then presented a framework that considers all the models in the hyperparameter optimization phase. 
Instead of the conventional approach of using one best model based on the lowest CV loss, we create an ensemble of feature importance using numerous good models in the training path for feature selection. Across the large-scale study in simulation settings and real-data experiments, we demonstrate the framework's ability to improve the feature selection's stability and power. 

In this work, we consider knockoff inference as a tool to perform feature selection due to its attractive statistical properties. However, our framework is general and applies to other feature selection approaches. Therefore, we consider the application of our framework to other feature selection methods as the immediate future goal. Further, we manually defined the search space for the hyperparameter optimization and thus, are likely to miss other optimal model parameters. We, therefore, consider integrating the presented framework with the recent advancements in Neural Architecture Search (NAS) \cite{baker2016designing,elsken2019neural} as the future direction for improving the stability and power of selected features.

\bibliographystyle{plainnat}
\bibliography{ref}

\appendix
\section{Appendix}
\label{appendxA}

\subsubsection{MNIST:}
The training architecture for MNIST dataset in feature importance instability analysis:
\begin{table}[!htbp]
  \centering
  \begin{tabular*}{\linewidth}[t]{c|c}
    \hline
Layer & Info (MLP for MNIST) \\
    \hline
    FC$_{1}$ & Linear (784 $\longrightarrow$ 512), ReLU \\
    \hline
    FC$_{2}$ & Linear (512 $\longrightarrow$ 512), ReLU \\
    \hline
    FC$_{3}$ & Linear (512 $\longrightarrow$ 10) \\
    \hline
  \end{tabular*} 
  \label{tab:mnistArch}
\end{table}

The network is trained for 200 epochs with a mini-batch size of 128 and a constant learning rate of 0.001 with the Adam optimizer. 

\subsubsection{CIFAR-10:} We used VGG-11 \cite{simonyan2014very} as the training architecture for training CIFAR-10 dataset. We followed the standard CIFAR-10 classification tutorial from PyTorch, which includes training with a scheduled learning rate with Cosine Annealing with a maximum number of iterations set to 200. SGD optimizer with a learning rate of 0.01, the momentum of 0.9, and weight decay of 5$e-4$, and a learning rate of 0.01 is used. The network is trained to 100 epochs. 

\subsubsection{Simulated dataset:} The training architecture for the simulated dataset ($p$ = 500), with single knockoff ($p$ = 500):
\begin{table}[!htbp]
  \centering
  \begin{tabular*}{\linewidth}[t]{c|c}
    \hline
Layer & Info (DeepLink for Simulated dataset) \\
    \hline
    PC$_{1}$ & PairWiseConnectedLayer (2 $\times$ $p$ $\longrightarrow$ $p$) \\
    \hline
    Drop$_{1}$ & Dropout (prob=0.5) \\
    \hline
    FC$_{1}$ & Linear ($p$ $\longrightarrow$ 25), ELU, L1-penalty ($\lambda$) \\
    \hline
    FC$_{2}$ & Linear (25 $\longrightarrow$ 25), ELU \\
    \hline
    FC$_{3}$ & Linear (25 $\longrightarrow$ 25), ELU \\
    \hline
    FC$_{\textrm{final}}$ & Linear (25 $\longrightarrow$ 1) \\
    \hline
  \end{tabular*} 
  \label{tab:simArch}
\end{table}

For depth = 1, we have (PC$_{1}$, Drop$_{1}$, FC$_{1}$, and FC$_{\textrm{final}}$) as our architecture. For depth = 2, we have (PC$_{1}$, Drop$_{1}$, FC$_{1}$, FC$_{2}$, and FC$_{\textrm{final}}$) as our architecture, and for depth = 3, we use all the layers in the table above. The resulting network is trained for 300 epochs with a mini-batch size of 32 and a constant learning rate of 0.001 with the Adam optimizer. 

\subsubsection{Proteomics dataset:} The training architecture for the Proteomics dataset with data $\mathbf{}{X}$ ($p$ = 7596) and covariates ($p_{\textrm{covariates}}$ = 10) and with $m_k = 5$ multiple knockoffs ($p$ = 37,980):
\begin{table}[!htbp]
  \centering
  \begin{tabular*}{\linewidth}[t]{c|c}
    \hline
Layer & Info (DeepLink for Proteomics dataset) \\
    \hline
    PC$_{1}$ & PairWiseConnectedLayer (6 $\times$ $p$ $\longrightarrow$ $p$) \\
    \hline
    Drop$_{1}$ & Dropout (prob=0.5) \\
    \hline
    FC$_{1}$ & Linear ($p$ $\longrightarrow$ 50), ELU, L1-penalty ($\lambda$) \\
    \hline
    FC$_{2}$ & Linear (50 $\longrightarrow$ 50), ELU \\
    \hline
    FC$_{3}$ & Linear (50 $\longrightarrow$ 50), ELU \\
    \hline
    FC$_{\textrm{concat}}$ & Concatenation(50, 10) [with covariates] \\
    \hline
    FC$_{\textrm{final}}$ & Linear (60 $\longrightarrow$ 1) \\
    \hline
  \end{tabular*} 
  \label{tab:simArch}
\end{table}

The network is trained for 200 epochs with a mini-batch size of 128 and a constant learning rate of 0.001 with the Adam optimizer. 

\subsubsection{Cancer \& Diabetes dataset:} The training architecture for the Cancer dataset with data $\mathbf{X}$ ($p$ = 30) and with $m_k = 5$ multiple knockoffs ($p$ = 150), and for the Diabetes dataset with data $\mathbf{X}$ ($p$ = 8) and with $m_k = 5$ multiple knockoffs ($p$ = 40):
\begin{table}[!htbp]
  \centering
  \begin{tabular*}{\linewidth}[t]{c|c}
    \hline
Layer & Info (DeepLink for Cancer and Diabetes dataset) \\
    \hline
    PC$_{1}$ & PairWiseConnectedLayer (6 $\times$ $p$ $\longrightarrow$ $p$) \\
    \hline
    Drop$_{1}$ & Dropout (prob=0.5) \\
    \hline
    FC$_{1}$ & Linear ($p$ $\longrightarrow$ 50), ELU, L1-penalty ($\lambda$) \\
    \hline
    FC$_{2}$ & Linear (50 $\longrightarrow$ 50), ELU \\
    \hline
    FC$_{3}$ & Linear (50 $\longrightarrow$ 50), ELU \\
    \hline
    FC$_{\textrm{final}}$ & Linear (50 $\longrightarrow$ 1) \\
    \hline
  \end{tabular*} 
  \label{tab:realArch}
\end{table}

For cancer, the network is trained for 100 epochs, and for the diabetes dataset, the network is trained for 50 epochs. Both datasets are trained with a mini-batch size of 128 and a constant learning rate of 0.001 with the Adam optimizer.

\section{Appendix}
\label{appendxB}
Formally, knockoffs aims to simultaneously test the hypotheses $H_j : X_j \perp\!\!\!\perp Y | X_{-j}$, for $j \in [p]$ \cite{candes2018panning}. If $\mathcal{H}_0 = \{j : H_j \}$ is the set of null hypotheses, the knockoff procedure selects a set of features $\hat{P}$ and provaly controls FDR when the distribution of $X$ is known:
\begin{equation}
    \textrm{FDR} = \mathbb{E}\bigg(\frac{|\hat{P} \cap \mathcal{H}_{0}|}{|\hat{P}|} \bigg) \leq q
\end{equation}
where $q \in [0, 1]$ is a pre-defined threshold for FDR, commonly known as the target FDR level. The knockoff procedure primarily contains four steps: constructing knockoffs, computing feature importance, computing feature statistics, and applying a knockoff filter to select significant features with FDR control. 
Here, we provide details for these steps:

\subsubsection{Constructing knockoffs:} For a given data $X \in \mathbb{R}^p$ with $p$ features, we define knockoffs $\Tilde{X} \in \mathbb{R}^p$ as random variable satisfying pairwise exchangeability condition and conditional independence, \textit{i.e.,}
\begin{equation}
[X, \Tilde{X}]_{\textrm{swap}(J)} \stackrel{d}{=} [X, \Tilde{X}] \quad \textrm{and} \quad \Tilde{X} \perp\!\!\!\perp Y | X
\end{equation}
for all $J \subset [p]$. This setup of creating single knockoff can have limitations in terms of detection threshold and stability which can be critical for real-world datasets. We thus consider multiple knockoffs for our real data experiments. For multiple knockoffs generation, we follow the efficient sequential conditional independent tuples (SCIT) algorithm from \cite{he2021identification}. 
The main steps of SCIT to yield a sequence of random variables obeying the exchangeability property is shown here:

\begin{algorithm}[H]
  \label{alg:algorithm1}
  \caption{Sequential Conditional Independent Tuples (Multiple Knockoffs)}
  j = 1 \\
  \textbf{do} \\
    Sample $\Tilde{X}_j^1, .., \Tilde{X}_j^M$ from $\mathcal{L}(X_j | X_{-j}, \Tilde{X}_{1:j-1}^{1}, .., \Tilde{X}_{1:j-1}^{M})$ \\
    j = j + 1 \\
  \textbf{while} $j \leq p$ 
\end{algorithm}
where $\mathcal{L}(X_j | X_{-j}, \Tilde{X}_{1:j-1}^{1}, .., \Tilde{X}_{1:j-1}^{M})$ is the conditional distribution of $X_j$ given $X_{-j}$, $\Tilde{X}^1_{1:j-1}, .., \Tilde{X}_{1:j-1}^{M})$.

\subsubsection{Calculating feature importance score:} The feature importance score $Z \in \mathbb{R}^{p + m*p}$ where $j \in [p]$ and $m \in [1,M]$ represents the importance of both original features and their knockoff copies, \textit{i.e., } $Z_j$ represents the importance of $X_j$, and $Z_{j+(m*p)}$ represents the importance of $\Tilde{X}^{m}_j$, $m^{\textrm{th}}$ copy in $M$ knockoffs. Obtaining feature importance score can be considered as the combination of mapping function $f(\cdot)$ and importance scoring function $z(\cdot)$, which can be represented as $Z = z(f([X,\Tilde{X}], Y)) \in \mathbb{R}^{p + m*p}$. 
Here, fitting function $f(\cdot)$ is represented by deep neural network, and interpretability score (e.g., Grads) as the corresponding scoring function $z(\cdot)$. 

\subsubsection{Calculating feature statistics:} Feature importance score of  original features $Z_j$ and knockoff features $Z_{j+(m*p)}$ are combined into a feature statistic $W_j = f_s (Z_j, Z_{j+(m*p)})$, where $f_s$ is an antisymmetric function such that $f_s(x, y) = - f_s(y,s)$. For single knockoff case, \textit{i.e.,} $m = 1$, if $Z$ are absolute coefficients (or absolute values of gradients for deep neural networks), we can set $f_s (Z_j, Z_{j+p}) = Z_j - Z_{j+p}$. This intuitively means that original features with higher feature importance score than its knockoff are likely to be an importance feature. For multiple knockoffs, we consider multiple-knockoff feature statistics from \cite{he2021identification}, where $W_j$ is defined as:
\begin{equation}
    \label{eq:calculatingW}
    W_j = \bigg(Z_j - \underset{1 \leq m \leq M}{\textrm{median}} Z_{j + (m*p)}\bigg) \mathbb{I}_{Z_j \geq \underset{1 \leq m \leq M}{\textrm{max}} Z_{j+p*m}}
\end{equation}
Here, $W_j$ essentially selects windows where the original feature has higher importance score than any of the $M$ knockoffs, and the gap with the median of knockoff importance score is above some threshold.

\subsubsection{Using knockoff filter for feature selection:} The final step in knockoff framework is feature section with FDR $\leq q$. For single knockoff, the knockoff threshold $T$ is defined as:
\begin{equation}
\label{eq:singleKnockoff}
    T = \textrm{min}\bigg\{t>0 : \frac{\#\{j: W_j \leq -t\} + 1}{\#\{j: W_j \geq t\}} \leq q \bigg\}
\end{equation}
and for multiple knockoffs, 
\begin{equation}
\label{eq:multipleKnockoff}
    T = \textrm{min}\bigg\{t>0 : \frac{\frac{1}{M} + \frac{1}{M}\#\{\kappa_j \geq 1, \tau_j \geq 1\}}{\#\{\kappa_j = 0, \tau_j = 0\}} \leq q \bigg\}
\end{equation}
where $`\#'$ denotes the number of elements in the set, 
$\kappa_j$
denotes the index of the original (denoted as 0) or knockoff feature that has the largest importance score, and 
$\tau_j$
denote the difference between the largest importance score and the median of the remaining importance scores. In both cases of single and multiple knockoffs, selecting features $\Tilde{P} = \{j: W_j \geq T \}$ guarantees FDR control at level $q$.


\section{Appendix}
Here, we present the algorithm for calculating the $m$ influential observations.

\begin{algorithm}[ht]
  \label{alg:algorithm2}
  \caption{Statistical leveraging for finding influential $Z$}
  $\mathbb{Z} = Z^{\textrm{all}} \in \mathbb{R}^{n \times (m*p)}$ $\leftarrow$ feature importance matrix across all the hyperparameters setting \par
  $H = \mathbb{Z}(\mathbb{Z}^{T}\mathbb{Z})^{-1}\mathbb{Z}^{T}$ $\leftarrow$ calculate Hat Matrix \par
  $H = UU^{T}$ $\leftarrow$ decomposition of $H$ where $U$ is any orthogonal basis for the column space of  $\mathbb{Z}$ \par
  $\mathbf{h}_{i} = \sum^{m*p}_{j = 1} U^{2}_{ij} = ||\mathbf{u}_{i}||^{2}$ $\leftarrow$ compute the leverage of the $i^{\textrm{th}}$ observation \par
  $S^{w}_{i} = \frac{\mathbf{h}_{i}}{\sum \mathbf{h}_{i}}$ $\leftarrow$ calculate the sampling weights for each observation \par
  $Z_{[1:m]} = \textrm{sample}(\mathbb{Z}, \textrm{size}=m, \textrm{weights}=S^{w})$
\end{algorithm}

\section{Appendix}
Here, we provide the details about the real-world dataset used in this study. 

\subsubsection{Proteomics} dataset is used to identify and monitor biomarkers by analyzing the proteins in the body fluids such as urine, serum, exhaled breath, and spinal fluid. Proteomics can also facilitate drug development by providing a comprehensive map of protein interactions associated with disease pathways. The dataset is acquired from Stanford Alzheimer's Disease Research Center (ADRC) consortium. 

\subsubsection{Breast Cancer} dataset is used to classify tumor cells into malignant (cancerous) or benign(non-cancerous). Cancer starts when cells in the breast begin to grow out of control. These cells usually form tumors that can be seen via X-ray or felt as lumps in the breast area. The dataset is acquired from the Kaggle platform and is freely available at this link https://www.kaggle.com/datasets/yasserh/breast-cancer-dataset.

\subsubsection{Pima Diabetes} dataset is used to diagnostically predict whether or not a patient has diabetes, based on certain diagnostic measurements included in the dataset. It comprises several medical predictor variables and one target variable, outcome. Predictor variables include the number of pregnancies the patient has had, their BMI, insulin level, age, etc. This dataset is also acquired from the Kaggle platform and is freely available at this link https://www.kaggle.com/datasets/uciml/pima-indians-diabetes-database.

\end{document}